\documentclass{article}
\usepackage{import}
\usepackage[a4paper, margin=3.5cm]{geometry}
\usepackage[utf8]{inputenc}
\usepackage{amsmath}
\usepackage{nicefrac}
\usepackage[T1]{fontenc}
\usepackage{amsfonts}
\usepackage{titlesec}
\titlespacing*{\section}
{0pt}{5.5ex plus 1ex minus .2ex}{4.3ex plus .2ex}
\titlespacing*{\subsection}
{0pt}{5.5ex plus 1ex minus .2ex}{4.3ex plus .2ex}

\usepackage{lipsum}
\usepackage{parskip}
\usepackage{enumitem}
\usepackage[svgnames]{xcolor} 
\usepackage[backref=page]{hyperref}
\hypersetup{
    colorlinks = true,
    linkcolor = DarkRed,
    citecolor = NavyBlue,
    pdfborder= 0 0 0,
    linktocpage=true,
}
\usepackage{mathtools}
\usepackage{float}
\usepackage{physics}
\usepackage[export]{adjustbox}
\usepackage{tabularx}
\usepackage{graphicx}
\usepackage{listings}
\usepackage{amsthm} 
\usepackage{multicol}
\usepackage{bbm} 
\usepackage{stmaryrd} 
\usepackage{bm}

\usepackage{booktabs}
\usepackage[square,numbers]{natbib}
\usepackage[font=small,labelfont=bf]{caption}
\usepackage[subrefformat=parens]{subcaption}
\usepackage[framemethod=TikZ]{mdframed}

\xdefinecolor{navyblue}{named}{Navy}
\xdefinecolor{orangered}{named}{OrangeRed}
\xdefinecolor{darkred}{named}{DarkRed}
\xdefinecolor{cadetblue}{named}{CadetBlue}

\newcommand{\LL}{\mathcal{L}}

\newcommand{\D}{\mathcal{D}}
\newcommand{\E}{\mathbb{E}}
\newcommand{\1}{\mathbbm{1}}
\newcommand{\N}{\mathcal{N}}
\newcommand{\Dn}{\mathcal{D}_n}

\newcommand{\p}[1]{#1_{\sim p}}
\newcommand{\q}[1]{#1_{\sim q}}

\newcommand{\Ap}{A_{\sim p}}
\newcommand{\Aq}{A_{\sim q}}
\newcommand{\Xp}{X_{\sim p}}

\newcommand{\vp}{v_{\sim p}}
\newcommand{\vq}{v_{\sim q}}

\renewcommand{\wp}{w_{\sim p}}
\newcommand{\wq}{w_{\sim q}}

\newcommand{\R}{\mathbb{R}}
\newcommand{\EMC}{\text{EMC}_{P, \epsilon}(\mathcal{T})}

\DeclareMathOperator*{\argmin}{argmin}

\newenvironment{proof_sketch}{

\proof
}{\endproof}

\newmdtheoremenv[
    middlelinecolor=red!30, middlelinewidth=1pt, backgroundcolor=red!5, roundcorner=0pt, innertopmargin=5pt, skipabove=\topskip
]{theorem}{Theorem}
\newmdtheoremenv[
    middlelinecolor=red!30, middlelinewidth=1pt, backgroundcolor=red!5, roundcorner=0pt, innertopmargin=5pt, skipabove=\topskip
]{proposition}[theorem]{Proposition}
\newmdtheoremenv[
    middlelinecolor=red!30, middlelinewidth=1pt, backgroundcolor=red!5, roundcorner=0pt, innertopmargin=5pt, skipabove=\topskip
]{lemma}[theorem]{Lemma}
\newmdtheoremenv[
    middlelinecolor=red!30, middlelinewidth=1pt, backgroundcolor=red!5, roundcorner=0pt, innertopmargin=5pt, skipabove=\topskip
]{hypothesis}[theorem]{Hypothesis}
\newmdtheoremenv[
    middlelinecolor=red!30, middlelinewidth=1pt, backgroundcolor=red!5, roundcorner=0pt, innertopmargin=5pt, skipabove=\topskip
]{assumption}{Assumption}

\newmdtheoremenv[
    middlelinecolor=red!30, middlelinewidth=1pt, backgroundcolor=red!5, roundcorner=0pt, innertopmargin=5pt, skipabove=\topskip
]{corollary}{Corollary}

\newmdtheoremenv[
    middlelinecolor=blue!30, middlelinewidth=1pt, backgroundcolor=blue!5, roundcorner=0pt, innertopmargin=5pt, skipabove=\topskip
]{definition}[theorem]{Definition}

\newmdtheoremenv[
    middlelinecolor=orange!30, middlelinewidth=1pt, backgroundcolor=orange!5, roundcorner=0pt, innertopmargin=5pt, skipabove=\topskip
]{remark}{Remark}
\newmdtheoremenv[
    middlelinecolor=orange!30, middlelinewidth=1pt, backgroundcolor=orange!5, roundcorner=0pt, innertopmargin=5pt, skipabove=\topskip
]{exercise}{Exercise}

\graphicspath{{./figures/}}
\DeclareGraphicsExtensions{.pdf,.png}

\title{\Large{Understanding the Double Descent Phenomenon in Deep Learning}}
\date{January 28, 2021}
\author{%
  \large{Marc Lafon}\footnote{Equal contribution, work done during the DAC master at Sorbonne Université, Paris, France, under the supervision of Prof. Gérard Biau.}\\
  \small{\texttt{lafon.ma.ml@gmail.com}}\\
   \and
  \large{Alexandre Thomas}\footnotemark[1]\\
  \small{\texttt{hi@alxthm.com}}
}

\begin{document}

\maketitle


\begin{abstract}
Combining empirical risk minimization with capacity control is a classical strategy in machine learning when trying to control the generalization gap and avoid overfitting, as the model class capacity gets larger. Yet, in modern deep learning practice, very large over-parameterized models (e.g. neural networks) are optimized to fit perfectly the training data and still obtain great generalization performance. Past the \emph{interpolation point}, increasing model complexity seems to actually lower the test error.

In this tutorial, we explain the concept of \emph{double descent} introduced by \cite{belkin_reconciling_2019}, and its mechanisms. Section \ref{section:1} sets the classical statistical learning framework and introduces the double descent phenomenon. By looking at a number of examples, section \ref{section:inductive biases} introduces \emph{inductive biases} that appear to have a key role in double descent by selecting, among the multiple interpolating solutions, a smooth empirical risk minimizer. Finally, section \ref{section:3} explores the double descent with two linear models, and gives other points of view from recent related works.
\end{abstract}


\setcounter{tocdepth}{3}
\tableofcontents{}
\newpage

\section{Generalization error : classical view and modern practice}
\label{section:1}

\subsection{Definitions and results from statistical learning}
\label{section:refreshers}
In statistical learning theory, the supervised learning problem consists of finding a good predictor $h_n: \mathbb{R}^d \rightarrow \{0, 1\}$, based on some training data $D_n$. The data is typically assumed to come from a certain distribution, i.e. $D_n = \{(X_1, Y_1), \dots, (X_n, Y_n)\}$ is a collection of $n$ i.i.d. copies of the random variables $(X, Y)$, taking values in $\mathbb{R}^d \times \{0, 1\}$ and following a data distribution $P(X, Y)$. We also restrict ourselves to a given class of predictors by choosing $h_n \in \mathcal{H}$.

\begin{definition}[True risk]
With $\ell(h(X), Y) = \mathbbm{1}_{(h(X) \neq Y)}$ the 0-1 loss, the \emph{true risk} (or \emph{true error}) of a predictor $h: \mathbb{R}^d \rightarrow \{0, 1\}$ is defined as
$$
L(h) = \mathbb{E}[\ell(h(X), Y)] = \mathbb{P}(h(X) \neq Y)
$$
The true risk is also called the \emph{expected risk} or the \emph{generalization error}.
\end{definition}

\begin{remark}
We choose in this section a classification setting, but a regression setting could be adopted as well, for instance with $Y$ and $h_n$ taking values in $\mathbb{R}$ (which we will sometimes do in the subsequent sections). In this case, the 0-1 loss is replaced by other loss functions, such as the squared error loss $\ell(\hat{y}, y) = (\hat{y} - y)^2$.
\end{remark}

In practice, the true distribution of $(X, Y)$ is unknown, so we have to resort to a proxy measure based on the available data.

\begin{definition}[Empirical risk]
The \emph{empirical risk} of a predictor $h: \mathbb{R}^d \rightarrow \mathbb{R}$ on a training set $D_n$ is defined as
$$
L_n(h) = \frac{1}{n} \sum_{i=1}^{n} \ell(h(X_i), Y_i)
$$
\end{definition}

\begin{definition}[Bayes risk]
A predictor $h^*: \mathbb{R}^d \rightarrow \{0, 1\}$ minimizing the true risk, i.e. verifying
$$
L(h^*) = \inf_{h: \mathbb{R}^d \rightarrow \{0, 1\}} L(h)
$$
is called a \emph{Bayes estimator}. Its risk $L^* = L(h^*)$ is called the \emph{Bayes risk}
\end{definition}

Using $D_n$, our objective is to find a predictor $h_n$ as close as possible to $h^*$.

\begin{definition}[Consistency]
A predictor $h_n$ is \emph{consistent} if 
$$
\mathbb{E} L(h_n) \underset{n \rightarrow \infty}{\rightarrow} L^*
$$
\end{definition}

The \emph{empirical risk minimization (ERM)} approach \cite{Vapnik1992} consists in choosing a predictor that minimizes the empirical risk on $D_n$ : $h_n^* \in \text{argmin}_{h \in \mathcal{H}} L_n(h)$. This is something that can be done or approximated in practice, thanks to a wide range of algorithms and optimization procedures, but it is also necessary to ensure that our predictor $h_n^*$ performs well in general and not only on training data. Depending on the chosen class of predictors $\mathcal{H}$, statistical learning theory can give us guarantees or insights to make sure $h_n^*$ generalizes well to unseen data.

\subsection{Classical view}
The gap between any predictor $h_n \in \mathcal{H}$ and $h^*$ can be decomposed as follows.
$$
L(h_n) - L^* = \underbrace{L(h_n) - \inf_{h \in \mathcal{H}} L(h)}_{\text{estimation error}} +  \underbrace{\inf_{h \in \mathcal{H}} L(h) - L^*}_{\text{approximation error}}
$$

\begin{remark}
In addition to the approximation error (approximating reality with a model) and estimation error (learning a model with finite data) which fits in the statistical learning framework and are the focus of this tutorial, there is actually another source of error, the \emph{optimization error}. This is the gap between the risk of the predictor returned by the optimization procedure and an empirical risk minimizer $h_n^*$.
\end{remark}

\begin{proposition}
\label{prop:classical-bound}
For any empirical risk minimizer $h_n^* \in \text{argmin}_{h \in \mathcal{H}} L_n(h)$, the estimation error verifies
$$
L(h_n^*) - \inf_{h \in \mathcal{H}} L(h) \leq 2 \sup_{h \in \mathcal{H}} |L_n(h) - L(h)|
$$
\end{proposition}

\begin{proof}
We have
$$
L(h_n^*) - \inf_{h \in \mathcal{H}} L(h)
\leq |L(h_n^*) - L_n(h_n^*)| + |L_n(h_n^*) - \inf_{h \in \mathcal{H}} L(h)|
$$

With 
$$|L(h_n^*) - L_n(h_n^*)| 
\leq \sup_{h \in \mathcal{H}} |L_n(h) - L(h)|$$
since $h_n^* \in \mathcal{H}$, and :
$$
|L_n(h_n^*) - \inf_{h \in \mathcal{H}} L(h)|
= |\inf_{h \in \mathcal{H}}L_n(h) - \inf_{h \in \mathcal{H}} L(h)|
\leq \sup_{h \in \mathcal{H}} |L_n(h) - L(h)|
$$
after separating the cases where $|\inf_{h \in \mathcal{H}}L_n(h) - \inf_{h \in \mathcal{H}} L(h)| > 0$ and $|\inf_{h \in \mathcal{H}}L_n(h) - \inf_{h \in \mathcal{H}} L(h)| < 0$.
\end{proof}

The classical machine learning strategy is to find the right $\mathcal{H}$ to keep both the approximation error and the estimation error low.
\begin{enumerate}
    \item When $\mathcal{H}$ is too small, no predictor $h \in \mathcal{H}$ is able to model the complexity of the data and to approach the Bayes risk. This is called \emph{underfitting}.
    \item When $\mathcal{H}$ is too large, the bound from proposition \ref{prop:classical-bound} (maximal generalization gap over $\mathcal{H}$) will increase, and the chosen empirical risk minimizer $h_n^*$ may generalize poorly despite having a low training error. This is called \emph{overfitting}.
\end{enumerate}

\vspace{.4cm}
\begin{remark}
Similarly, the expected error can also be decomposed into a bias term due to model mis-specification and a variance term due to random noise being modeled by $h_n^*$. This is the \emph{bias-variance} trade-off, and is also highly dependent on the capacity of $\mathcal{H}$, the chosen class of predictors.
\end{remark}

\begin{exercise} [Bias-Variance decomposition]
Assume that $Y = h(X) + \epsilon$, with $\mathbb{E}[\epsilon] = 0, Var(\epsilon) = \sigma^2$. Show that, for any $x \in \mathbb{R}^d$, the expected error of a predictor $h_n$ obtained with the random dataset $D_n$ is :
$$
\mathbb{E}[(Y - h_n(X))^2 | X=x] = (h(x) - \mathbb{E}h_n(x))^2 + \mathbb{E}[(\mathbb{E}h_n(x) - h_n(x))^2] + \sigma^2
$$
\end{exercise}

In order to ensure a consistent estimator $h_n$, we can control $\mathcal{H}$ explicitly e.g. by choosing the number of features used in a linear classifier, or the number of layers of a neural network.

\vspace{.4cm}
\begin{theorem}[Vapnik-Chervonenkis inequality]
For any data distribution $P(X,Y)$, by using $V_{\mathcal{H}}$ the VC-dimension of the class $\mathcal{H}$ as a measure of the class complexity, one has
$$
\mathbb{E} \sup_{h\in\mathcal{H}} |L_n(h) - L(h)|
    \leq 4 \sqrt{\frac{V_{\mathcal{H}} \log(n+1)}{n}}
$$
\end{theorem}

A complete introduction to Vapnik-Chervonenkis theory is outside the scope of this tutorial, but $V_{\mathcal{H}}$ can be defined as the cardinality of the largest set of points that can be shattered, i.e. there is at least one $h \in \mathcal{H}$ that can assign all possible labels to the set. Combining this result with proposition \ref{prop:classical-bound} gives a useful bound on the generalization error for a number of model classes. For instance, if $\mathcal{H}$ is a class of linear classifiers using $d$ features (potentially non-linear transformations of input $x$), then we have : $V_{\mathcal{H}} \leq d+1$.

Other measures of the richness of the model class $\mathcal{H}$ also exist, such as the \emph{Rademacher complexity}, and can be useful in situations where $V_{\mathcal{H}} = +\infty$, or in regression settings.

\subsection{Modern practice}
\begin{figure}[ht]
    \centering
    \includegraphics[width=\linewidth]{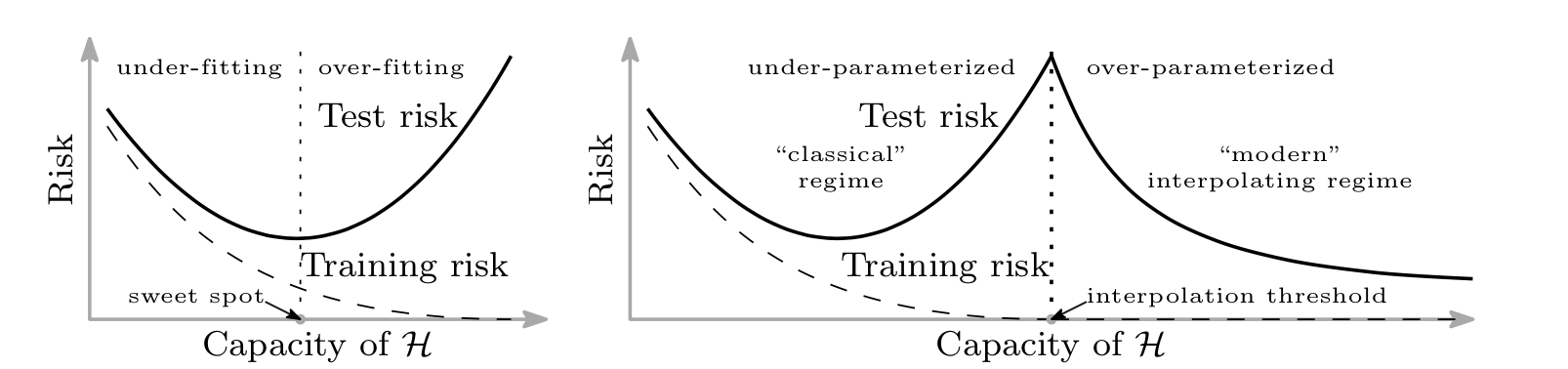}
    \caption{The \emph{classical risk curve} arising from the bias-variance trade-off and the \emph{double descent risk curve} with the observed modern interpolation regime. Taken from \cite{Belkin2019}}       
    \label{fig:double_descent_schema}
\end{figure}

Following results from section \ref{section:refreshers}, a widely adopted view is that, after a certain threshold, “larger models are worse” as they will overfit and generalize poorly. Yet, in modern machine learning practice, very large models with enough parameters to reach almost zero training error are frequently used. Such models are able to fit almost perfectly (i.e. \emph{interpolate}) the training data and still generalize well, actually performing better than smaller models (e.g. to classify 1.2M examples, AlexNet had 60M parameters and VGG-16 and VGG-19 both exceeded 100M parameters \cite{Canziani2016}). Understanding generalization of overparameterized models in modern deep learning is an active field of research, and we focus on the \emph{double descent} phenomenon, first demonstrated by \cite{Belkin2019} and illustrated in Figure \ref{fig:double_descent_schema}.

\vspace{.4cm}

For simpler class of models, classical statistical learning guarantee that the test risk decreases when the class of models gets more complex, until a point where the bounds do not control the risk anymore. However it seems that, beyond a certain threshold, increasing the capacity of the models actually decreases the test risk again. This is the “modern” interpolating regime, with overparameterized models. As this phenomenon depends not only on the class of predictors $\mathcal{H}$, but also on the training algorithm and regularization techniques, we define a \emph{training procedure} $\mathcal{T}$ to be any procedure that takes as input a dataset $D_n$ and outputs a classifier $h_n$, i.e. $h_n = \mathcal{T}(D_n) \in \mathcal{H}$. We can now make an informal hypothesis, after defining the notion of \emph{effective model complexity} (from \cite{Nakkiran2019}).

\vspace{.4cm}
\begin{definition}[Effective Model Complexity]
The \emph{Effective Model Complexity (EMC)} of a training procedure $\mathcal{T}$, w.r.t. distribution $P(X,Y)$, is the maximum number of samples $n$ on which $\mathcal{T}$ achieves on average $\approx 0$ training error. That is, for $\epsilon > 0$ :
$$
\EMC = \max\{n \in \mathbb{N} | \mathbb{E} L(h_n) \leq \epsilon\}
$$
\end{definition}

\vspace{.4cm}
\begin{hypothesis}[Generalized Double Descent hypothesis, informal]
For any data distribution $P(X,Y)$, neural-network-based training procedure $\mathcal{T}$, and small $\epsilon > 0$, if we consider the task of predicting labels based on $n$ samples from $P$ then, as illustrated on figure \ref{fig:double_descent_schema}:
\begin{itemize}
    \item \emph{Under-parameterized regime}. If $\EMC$ is sufficiently smaller than n, any perturbation of $\mathcal{T}$ that increases its effective complexity will decrease the test error.
    \item \emph{Critically parameterized regime}. If $\EMC \approx n$, then a perturbation of $\mathcal{T}$ that increases its effective complexity might decrease or increase the test error.
    \item \emph{Over-parameterized regime}. If $\EMC$ is sufficiently larger than n, any perturbation of $\mathcal{T}$ that increases its effective complexity will decrease the test error.
\end{itemize}
\end{hypothesis}

\begin{figure}[ht]
    \begin{subfigure}{0.49 \textwidth}
        \centering
        \includegraphics[width=\linewidth]{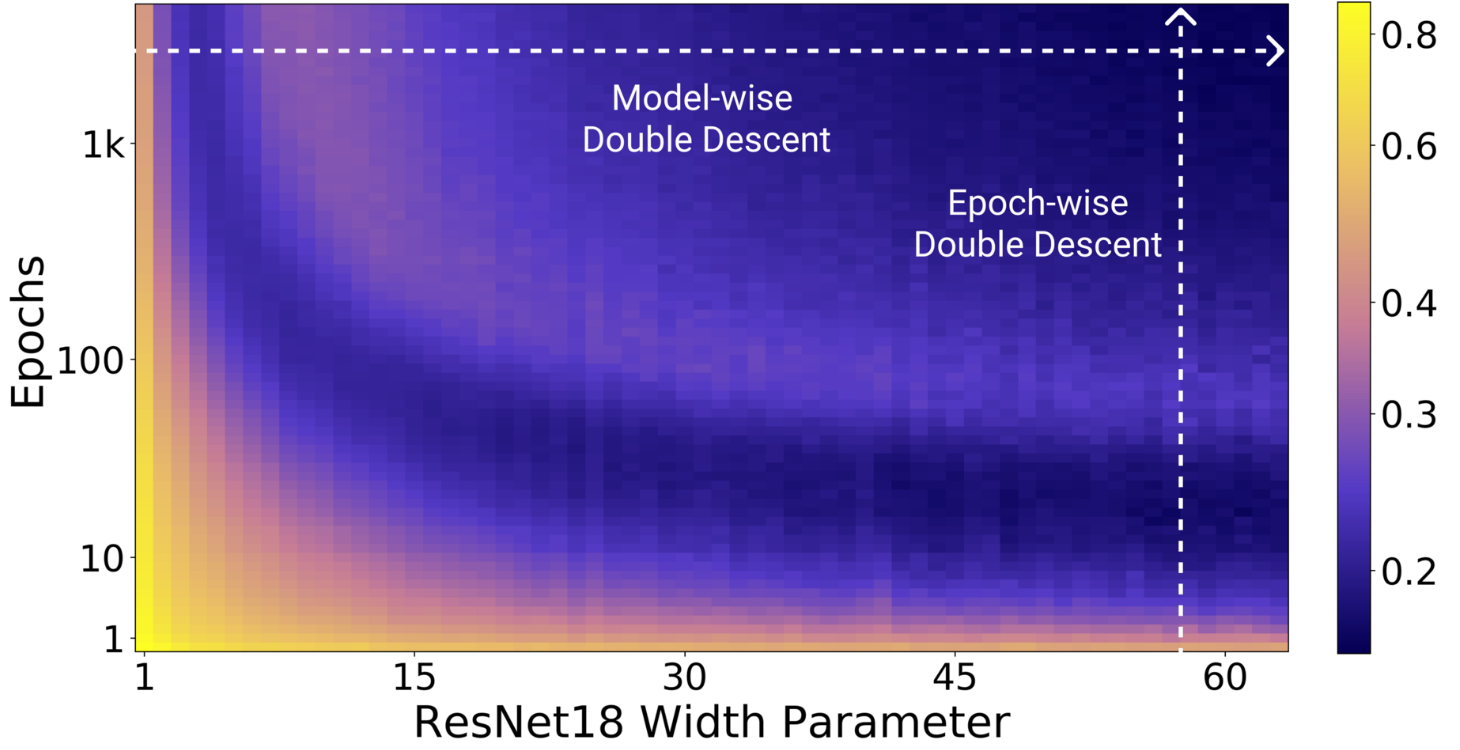}
        \caption{Test error as a function of model size and train epochs}
    \end{subfigure}
    \begin{subfigure}{0.49 \textwidth}
        \centering
        \includegraphics[width=\linewidth]{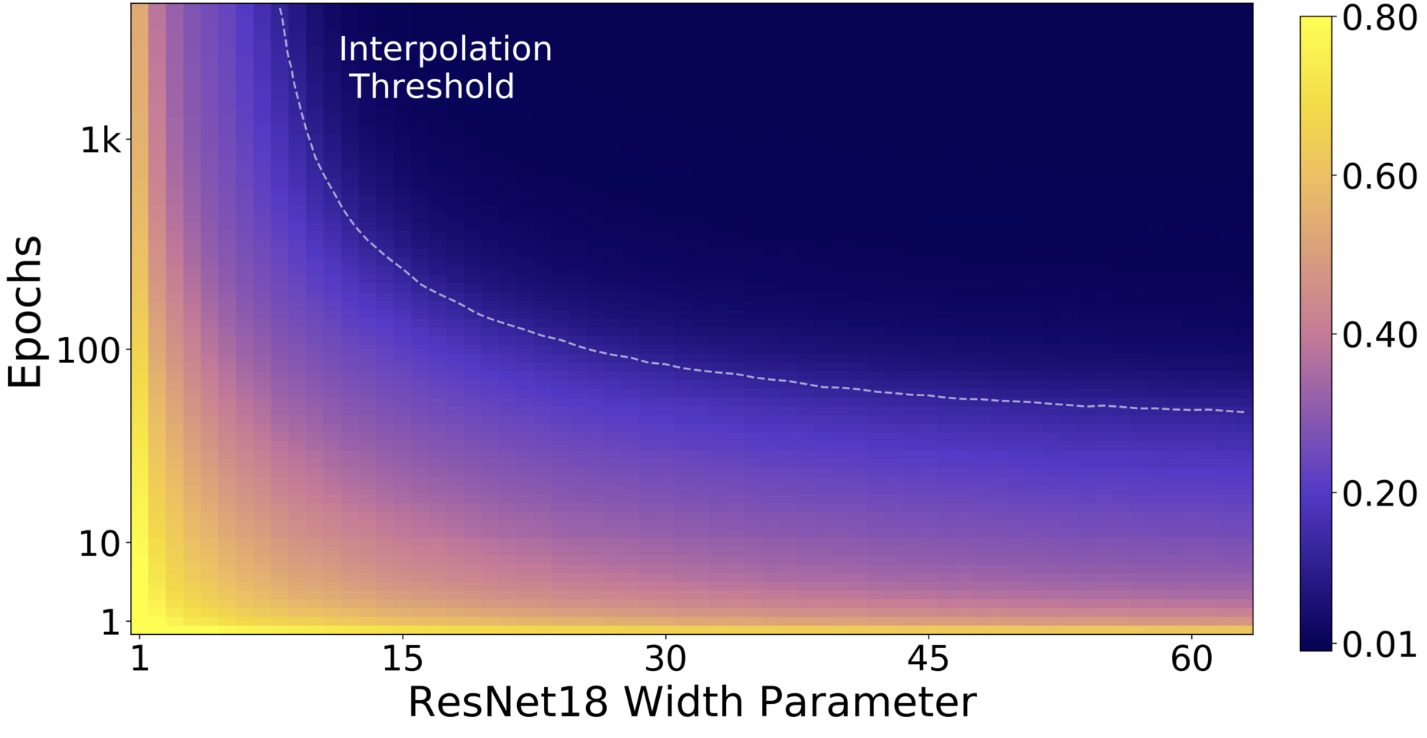}
        \caption{Train error of the corresponding models}
    \end{subfigure}
    \caption{All models are Resnet18s trained on CIFAR-10 with 15\% label noise (training labels artificially made incorrect), data-augmentation, and Adam for up to 4K epochs. Taken from \cite{Nakkiran2019}}
    \label{fig:double_descent_openai}
\end{figure}

Empirically, this definition of effective model capacity translates into multiple axis along which the double descent can be observed : \emph{epoch-wise}, \emph{model-wise} (e.g. increasing the width of convolutional layers or the embedding dimension of transformers) and even with regularization, by decreasing weight decay. Figure \ref{fig:double_descent_openai} illustrates this.

\vspace{.5cm}

\newpage
\section{Inductive biases}
\label{section:inductive biases}
%
In the supervised learning problem, the model needs to generalize patterns observed in the training data to unseen situations. In that sense, the learning procedure has to use mechanisms similar to inductive reasoning. As there are generally many possible generalizable solutions, \cite{Mitchell80} advocated the need for inductive biases in learning generalization. Inductive biases are assumptions made in order to prioritized one solution over another both exhibiting the same performance on the training data. For example, a common inductive bias is the Occam's razor principle stating that in case of equally good solutions the “simplest” one should be preferred. Another form of inductive bias is to incorporate some form of prior knowledge about the structure of the data, its generation process or to constrain the model to respect specific properties.

In the under-parameterized regime, regularization can be used for capacity control and is a form of inductive bias. One common choice is to search for small norm solutions, e.g. adding a penalty term, the $L_2$ norm of the weights vector. This is known as Tikhonov regularization in the linear regression setting (also known as Ridge regression in this case).

In the over-parameterized regime, as the complexity of $\mathcal{H}$ and the EMC increases, the number of interpolating solutions (i.e. achieving almost zero training error) increases and the question of the selection of a particular element in $\text{argmin}_{h \in \mathcal{H}} L_n(h)$ is crucial. Inductive biases, explicit or implicit, are a way to find predictors that generalize well.

\subsection{Explicit inductive biases}
Several common inductive biases can be used to observe a model-wise double descent \cite{belkin_reconciling_2019} (e.g. as the number of parameters $N$ increases).

\paragraph{Least Norm} For the model class of Random Fourier Features (defined in section \ref{section:RFF}), by choosing explicitly the minimum norm linear regression in the feature space. This bias towards the choice of parameters of minimum norm is common to a lot of machine learning model. For example, the ridge regression induce a constraint on the $L_2$ norm of the solution and the lasso regression on the $L_1$ norm. We can also see the support vector machine (SVM) as a way of inducing a least norm bias because maximizing the margin is equivalent to minimizing the norm of the parameter under the constraint that all points are well classified.

\paragraph{Model architecture} Another way of inducing a bias is by choosing a particular class of functions that we think is well suited for our problem. The authors in \cite{battaglia_relational_2018} discuss different type of inductive bias considered by different type of neural network architectures. Working with images it is better to use a convolutional neural network (CNN) as it can induce translational equivariance, whereas the recurrent neural network (RNN) is better suited to capture long-term dependencies in a sequence data. Using a naive Bayes classifier is of great utility if we know that the features are independent, etc.

\paragraph{Ensembling} Random forest models use yet another type of inductive bias. By averaging potentially non-smooth interpolating trees, the interpolating solution has a higher degree of smoothness and generalizes better than any individual interpolating tree.


\subsection{Implicit Bias of gradient descent}
Gradient descent is a widely used optimization procedure in machine learning, and has been observed to converge on solutions that generalize surprisingly well, thanks to an implicit inductive bias.

We recall that the gradient descent update rule for parameter $w$ using a loss function $\LL$ is the following (where $\eta >0$ is the step size):
\begin{equation*}
    \begin{aligned}
    w_{k+1} = w_k - \eta \nabla \LL(w)
    \end{aligned}
\end{equation*}

\subsubsection{Gradient descent in under-determined least squares problem}
\label{section:gd_least_square}
Consider a non-random dataset $\{(x_i, y_i)\}_{i=1}^n$, with $(x_i, y_i) \in \R^d\times\R$, for $i \in \{1, \dots ,n\}$ and let $\bm{X}\in \R^{n\times d}$ be the matrix which rows are the $x_i^T$ and $y \in \R^{n}$ the column vector which elements are the $y_i$.
We consider the linear least squares:

\begin{equation}
    \label{eqn:leastsquare}
    \min_{w\in \R^d} \LL(w) = \min_{w\in \R^d} \frac{1}{2}\norm{\bm{X} w - y}^2
\end{equation}

We will study the property of the solution found using gradient descent.

\begin{definition}[\emph{Moore-Penrose pseudo-inverse}]
\label{def:pseudo_inv}
Let $\bm{A} \in \R^{ n\times d}$ be a matrix, the Moore-Penrose pseudo-inverse is the only matrix $\bm{A}^{+}$ satisfying the following properties:
\setlength{\multicolsep}{3.pt}
\begin{multicols}{2}
\begin{enumerate}[label=(\roman*), nosep, align=parleft, labelsep=10mm, leftmargin = 3cm]
\item $\bm{A} \bm{A}^+ \bm{A} = \bm{A}$
\item $\bm{A}^+ \bm{A} \bm{A}^+ = \bm{A}^+$
\item $(\bm{A}^+\bm{A})^T = \bm{A}^+\bm{A}$
\item $(\bm{A}\bm{A}^+)^T = \bm{A}\bm{A}^+$
\end{enumerate}
\end{multicols}

Furthermore, if $\rank(\bm{A})=min(n,d)$ then $\bm{A}^+$ has a simple algebraic expression:
\begin{itemize}[label=-,nosep]
\item If $n<d$, then $\rank(\bm{A})=n$ and $\bm{A}^+=\bm{A}^T(\bm{A}\bm{A}^T)^{-1}$
\item If $d<n$, then $\rank(\bm{A})=d$ and $\bm{A}^+=(\bm{A}^T\bm{A})^{-1}\bm{A}^T$
\item If $d=n$, then $\bm{A}$ is invertible and $\bm{A}^+=\bm{A}^{-1}$
\end{itemize}
\end{definition}

\begin{lemma}
\label{lemma:psdinv_prop}
For a matrix $\bm{A} \in \R^{ n\times d}$, $Im(I\text{-}\bm{A}^+\bm{A})=Ker(\bm{A})$, $Ker(\bm{A}^+)=Ker(\bm{A}^T)$ and $Im(\bm{A}^+)=Im(\bm{A})$.
\end{lemma}

\begin{proof} Left as an exercise.
\end{proof}

\begin{theorem}
\label{thm:ls_solutions}
The set of solutions $\mathcal{S}_{LS}$ of the least square problem (i.e. minimizing (\ref{eqn:leastsquare}))  is exactly:

\centerline{$\mathcal{S}_{LS} = \{\bm{X}^+y + (\bm{I}\text{-}\bm{X}^+\bm{X})u, u\in \R^d\}$}
\end{theorem}


\begin{proof_sketch}
Writing $\bm{X} w - y = \bm{X} w - \bm{X}\bm{X}^+y - (\bm{I}-\bm{X}\bm{X}^+)y$ prove using pseudo-inverse properties that $\bm{X} w - \bm{X}\bm{X}^+y$ and $(\bm{I}-\bm{X}\bm{X}^+)y$ are orthogonal. Then using the Pythagorean theorem show that $\norm{\bm{X} w - y}^2 \geq  \norm{(\bm{I}-\bm{X}\bm{X}^+)y}^2$, this inequality being an equality if and only if $\bm{X}w=\bm{X}\bm{X}^+y$. Then $\bm{X}^+y$ is one solution of (\ref{eqn:leastsquare}) and by Lemma \ref{lemma:psdinv_prop} we can conclude that $\{\bm{X}^+y + (\bm{I}-\bm{X}^+\bm{X})u, u\in\R^d\}$, is the set of solutions.
\end{proof_sketch}


\begin{remark}
Depending on the $\rank$ of $\bm{X}$, the set of solutions $\mathcal{S}_{LS}$ will differ depending on the expression of $\bm{X}^+$:
\begin{itemize}[label=-]
    \item If $n<d$ and $\rank(\bm{X})=n$, then $\bm{X}^+=\bm{X}^T(\bm{X}\bm{X}^T)^{-1}$: $\mathcal{S}_{LS} = \{\bm{X}^T(\bm{X}\bm{X}^T)^{-1}y + (\bm{I}-\bm{X}^T(\bm{X}\bm{X}^T)^{-1}\bm{X})u, u\in\R^d\}$
    \item If $d<n$ and $\rank(\bm{X})=d$, then $\bm{X}^+=(\bm{X}^T\bm{X})^{-1}\bm{X}^T$: $\mathcal{S}_{LS} = \{\bm{X}^T(\bm{X}\bm{X}^T)^{-1}y\}$
    \item If $d=n$ and $\bm{X}$ is invertible, then $\bm{X}^+=\bm{X}^{-1}$: $\mathcal{S}_{LS} = \{\bm{X}^{-1}y\}$
    
\end{itemize}
\end{remark}

\begin{proposition}
\label{prop:smalestnorm}
Assuming that $\bm{X}$ has $\rank n$ and $n<d$, the least square problem (\ref{eqn:leastsquare}) has infinitely many solutions and  $\bm{X}^+y = \bm{X}^T(\bm{X}\bm{X}^T)^{-1}y$ is the minimum euclidean norm solution.
\end{proposition}

\begin{proof}
From the previous remark, we know that 
$$\ \mathcal{S}_{LS} = \{\bm{X}^T(\bm{X}\bm{X}^T)^{-1}y + (\bm{I}-\bm{X}^T(\bm{X}\bm{X}^T)^{-1}\bm{X})u, u\in\R^d\}$$

For arbitrary $u\in \R^d$,
\begin{equation*}
    \begin{aligned}
(\bm{X}^+y)^T(\bm{I}-\bm{X}^+\bm{X})u \overset{\mathrm{(ii)}}{=} (\bm{X}^+\bm{X}\bm{X}^+y)^T(\bm{I}-\bm{X}^+\bm{X})u
                  &= (\bm{X}^+y)^T(\bm{X}^+\bm{X})^T(\bm{I}-\bm{X}^+\bm{X})u\\
                  &\overset{\mathrm{(iii)}}{=} (\bm{X}^+y)^T\bm{X}^+\bm{X}(\bm{I}-\bm{X}^+\bm{X})u\\
                  &= (\bm{X}^+y)^T\bm{X}^+(\bm{X}-\bm{X}\bm{X}^+\bm{X})u \overset{\mathrm{(i)}}{=} 0
    \end{aligned}
\end{equation*}
using $(i)$, $(ii)$ and $(iii)$ from Definition \ref{def:pseudo_inv}. Thus, $(\bm{X}^+y)$ and $(\bm{I}-\bm{X}^+\bm{X})u$ are orthogonal $\forall u \in \R^d$, and applying the Pythagorean theorem gives:
\begin{equation*}
    \begin{aligned}
\norm{(\bm{X}^+y)+(\bm{I}-\bm{X}^+\bm{X})u}^2 &= \norm{(\bm{X}^+y)}^2+\norm{(\bm{I}-\bm{X}^+\bm{X})u}^2 \geq \norm{(\bm{X}^+y)}^2 \\
    \end{aligned}
\end{equation*}
\end{proof}

\begin{theorem}
\label{thm:gd_ls}
If the linear least square problem (\ref{eqn:leastsquare}) 
is under-determined, i.e. $(n<d)$ and $\rank(\bm{X})=n$, using gradient descent with a fixed learning rate $0<\eta<\frac{1}{\sigma_{max}(\bm{X})}$, where $\sigma_{max}(\bm{X})$ is the largest eigenvalue of $\bm{X}$, from an initial point $w_0\in Im(\bm{X}^T)$ will converge to the minimum norm solution of (\ref{eqn:leastsquare}).
\end{theorem}

\begin{proof}
As $\bm{X}$ is assumed to be of row rank $n$, we can write its singular value decomposition as :
\[
\bm{X} = \bm{U} \bm{\Sigma} \bm{V}^T = \bm{U} 
\begin{bmatrix}\bm{\Sigma}_1 & 0 \end{bmatrix} \begin{bmatrix}\bm{V}_1^T \\ \bm{V}_2^T \end{bmatrix}
\]
where $\bm{U}\in \R^{n\times n}$ and $\bm{V}\in \R^{d\times d}$ are orthogonal matrices, $\bm{\Sigma} \in \R^{n\times d}$ is a rectangular diagonal matrix and $\bm{\Sigma}_1 \in \R^{n\times n}$ is a diagonal matrix. The  minimum norm solution $w^*$ can be rewritten as :
\[
w^* = \bm{X}^T(\bm{X}\bm{X}^T)^{-1}y = \bm{V}_1 \bm{\Sigma}_1^{-1}\bm{U}^Ty
\]
The gradient descent update rule is the following (where $\eta >0$ is the step size):
\begin{equation*}
    \begin{aligned}
    w_{k+1} = w_k - \eta \nabla \LL(w)
            = w_k - \eta \bm{X}^T(\bm{X} w_k - y)
            = (\bm{I}-\eta \bm{X}^T\bm{X})w_k + \eta \bm{X}^Ty
    \end{aligned}
\end{equation*}

Then, by induction, we have :
\[
w_{k} = (\bm{I}-\eta \bm{X}^T\bm{X})^k w_0 + \eta \sum_{l=0}^{k-1} (\bm{I}-\eta \bm{X}^T\bm{X})^l \bm{X}^Ty\\
\]

Using the singular value decomposition of $\bm{X}$, we can see that $\bm{X}^T\bm{X} = \bm{V} \bm{\Sigma}^T \bm{\Sigma} \bm{V}^T$. Furthermore, as $\bm{V}$ is orthogonal, $\bm{V}^T\bm{V}=\bm{I}$.\\
Then, the gradient descent iterate at step $k$ can be written:
\begin{equation*}
    \begin{aligned}
    w_k &= \bm{V}(\bm{I}-\eta\bm{\Sigma}^T\bm{\Sigma})^k \bm{V}^T w_0 + \eta \bm{V} \Big(\sum_{l=0}^{k-1} (\bm{I} - \eta \bm{\Sigma}^T \bm{\Sigma})^l \bm{\Sigma}^T \Big) \bm{U}^Ty\\
    &= \bm{V}
    \begin{bmatrix}
    (\bm{I}-\eta\bm{\Sigma}_1^2)^k & 0 \\ 
    0 & \bm{I} 
    \end{bmatrix}
    \bm{V}^T w_0 + \eta \bm{V} \Big(\sum_{l=0}^{k-1} 
    \begin{bmatrix}
    (\bm{I}-\eta\bm{\Sigma}_1^2)^l \bm{\Sigma}_1 \\ 
    0 
    \end{bmatrix}
    \Big) \bm{U}^Ty
    \end{aligned}
\end{equation*}
By choosing $0<\eta<\nicefrac{1}{\sigma_{max}(\bm{\Sigma}_1)}$ with $\sigma_{max}(\bm{\Sigma}_1)$ the largest eigenvalue of $\bm{\Sigma}_1$, we guarantee that the eigenvalues of $\bm{I}-\eta\bm{\Sigma}^T \bm{\Sigma}$ are all strictly less than 1. 
Then :
\[
\bm{V}\begin{bmatrix}
    (\bm{I}-\eta\bm{\Sigma}_1^2)^k & 0 \\ 
    0 & \bm{I} 
 \end{bmatrix} \bm{V}^T w_0 \xrightarrow[k\rightarrow \infty]{} \bm{V}\begin{bmatrix}
    0 & 0 \\ 
    0 & \bm{I} 
 \end{bmatrix} \bm{V}^T w_0 = \bm{V}_2 \bm{V}_2^T w_0
\]
and
\[
\eta \sum_{l=0}^{k-1} 
    \begin{bmatrix}
    (\bm{I}-\eta\bm{\Sigma}_1^2)^l \bm{\Sigma}_1 \\ 
    0 
    \end{bmatrix} \xrightarrow[k\rightarrow \infty]{} 
    \eta  
    \begin{bmatrix}
    \sum_{l=0}^{\infty}(\bm{I}-\eta\bm{\Sigma}_1^2)^l \bm{\Sigma}_1 \\ 
    0 
    \end{bmatrix} = \begin{bmatrix}
    \eta (\bm{I}- \bm{I} + \eta \bm{\Sigma}_1^2)^{-1}\bm{\Sigma}_1\\ 
    0 
    \end{bmatrix} =  \begin{bmatrix}
    \bm{\Sigma}_1^{-1}\\ 
    0 
    \end{bmatrix}
\]

Finally, noting $w_\infty$ the limit of gradient descent iterates we have in the limit :
\begin{equation*}
    \begin{aligned}
    w_{\infty} &= \bm{V}_2 \bm{V}_2^T w_0 + \bm{V}_1 \bm{\Sigma}_1^{-1} \bm{U}^Ty
              &= \bm{V}_2 \bm{V}_2^T w_0 + \bm{X}^T(\bm{X}\bm{X}^T)^{-1}y
              &= \bm{V}_2 \bm{V}_2^T w_0 + w^*
    \end{aligned}
\end{equation*}

Because $w_0$ is in the range of $\bm{X}^T$, we can write $w_0 = \bm{X}^T z$ for some $z \in \R^n$.

\begin{equation*}
    \begin{aligned}
    \bm{V}_2 \bm{V}_2^T w_0 = \bm{V}\begin{bmatrix}
                         0 & 0 \\ 
                         0 & \bm{I} 
                      \end{bmatrix} \bm{V}^T \bm{X}^Tz
                  &= \bm{V}\begin{bmatrix}
                         0 & 0 \\ 
                         0 & \bm{I} 
                      \end{bmatrix} \bm{V}^T \bm{V} \bm{\Sigma}^T \bm{U}^Tz
                  &= \bm{V}\begin{bmatrix}
                         0 & 0 \\ 
                         0 & \bm{I} 
                      \end{bmatrix} \begin{bmatrix}\bm{\Sigma}_1\\ 0 \end{bmatrix} \bm{U}^T=0
    \end{aligned}
\end{equation*}
Therefore gradient descent will converge to the minimum norm solution.
\end{proof}

\subsubsection{Gradient descent on separable data}

In this section we are concerned with the effect of using gradient descent on a classification problem on a linearly separable dataset and using a smooth (we will explain in what sens), strictly decreasing and non-negative surrogate loss function. For the sake of clarity, we will prove the results using the exponential loss function $\ell:x\mapsto e^{-x}$ but the results will be expressed for the more general case.

\vspace{.4cm}
\begin{definition}[\emph{Linearly separable dataset}]
A dataset $\Dn = \{(x_i, y_i)\}_{i=1}^{n}$ where $\forall i \in [\![ 1, n]\!], (x_i, y_i) \in \R^d\times\{-1,1\}$ is linearly separable if $\exists\ w_*$ such that $\forall i: y_i w_*^T x_i > 0$.
\end{definition}

The results of this section hold assuming the considered loss functions respect the following properties : 

\begin{assumption}
The loss function $\ell$ is positive, differentiable, monotonically decreasing to zero, (i.e. $\ell(u)>0$, $\ell'(u)<0$, $\lim_{u \xrightarrow{}\infty}\ell(u)=\lim_{u \xrightarrow{}\infty}\ell'(u)=0$) and $\lim_{u \xrightarrow{}-\infty}\ell'(u)\neq0$.
\end{assumption}

\begin{assumption}[$\beta$-Smoothness]
The gradient of $\ell$ is $\beta$-Lipschitz: 
\[
\ \ \forall u,v \in \R, \ \ \norm{\nabla \ell(u) - \nabla \ell(v)}\leq \beta \norm{u-v}.
\]
\end{assumption}

\begin{assumption}[Tight Exponential tail]
Generally speaking a function $f:\R \mapsto \R$ is said to have a \emph{tight exponential tail} if there exist positive constants c, a, $\mu_1$, $\mu_2$ and $u_0$ such that:
\[
\forall u >u_0,\ (1-e^{-\mu_1u})\leq c\ f(u) e^{au} \leq (1+e^{-\mu_2u}).
\]
In our case we will say that a differentiable loss function $\ell$ has a \emph{tight exponential tail} when its negative derivative $-\ell'$ has a tight exponential tail.
\end{assumption}

\vspace{.5cm}
\begin{figure}[ht]
\centering
    \begin{subfigure}[b]{0.45\textwidth}
        \centering
        \includegraphics[width=\textwidth]{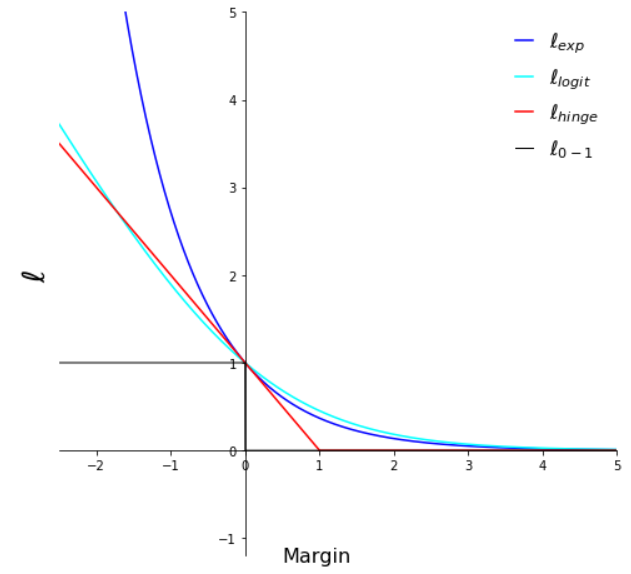}
        \caption{Losses}
    \end{subfigure}
    \begin{subfigure}[b]{0.45\textwidth}
        \centering
        \includegraphics[width=\textwidth]{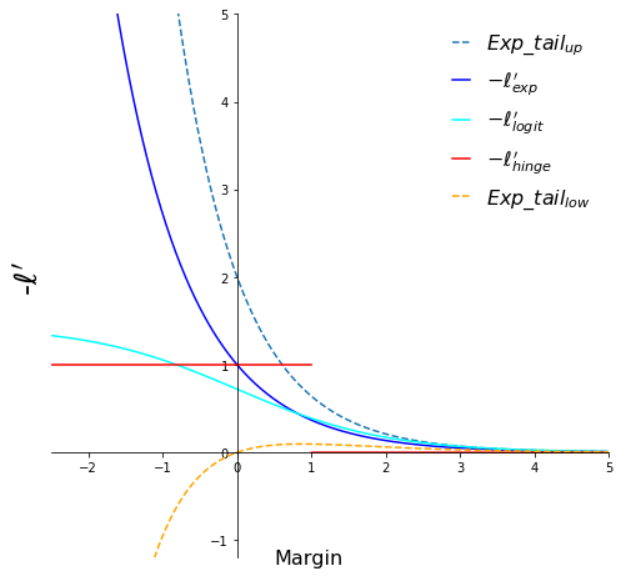}
        \caption{Negative Derivative Losses}
    \end{subfigure}
    \caption{Illustration of tight exponential tail property for different common loss functions. We can see that both exponential and logistic loss functions has a tight exponential tail. The hinge loss and 0-1 loss functions have been displayed for reference only.}
    \label{fig:tight_exp_tail}
\end{figure}

We consider the following classification problem:
\[
    \min_{w\in \R^d} \LL(w) = \min_{w\in \R^d} \sum_{i=1}^{n}\ell(y_i w^T x_i)
\]
where $\forall i \in [\![ 1, n]\!], (x_i, y_i) \in \R^d\times\{-1,1\}$ and $\ell:\R \mapsto \R^*_+$ is a surrogate loss function of the $0$-$1$ loss.

We will study the behavior of the solution found by gradient descent using a fixed learning rate $\eta$:
\begin{equation}
    w_{t+1} = w_{t} - \eta \nabla \LL(w_t) = w_{t} - \eta \sum_{i=1}^{n}\ell'(y_i w_t^T x_i)y_i x_i
\end{equation}

\begin{lemma}
\label{lemma:exploding_norm}
Let $\D = \{(x_i, y_i)\}_{i=1}^{n}$ be a linearly separable dataset where $\forall i \in [\![ 1, n]\!]$, $(x_i, y_i) \in \R^d\times\{-1,1\}$ and $\ell:\R \mapsto \R^*_+$ a loss function under assumptions 1 and 2. Let $w_t$ be the iterates of gradient descent using learning rate $0<\eta<\frac{2}{\beta\sigma^2_{max}(X)}$ and any starting point $w_0$. Then we have:
\begin{enumerate}[label=(\arabic*),nosep, align=parleft, labelsep=10mm, leftmargin = 3cm]
    \item $\lim_{t \xrightarrow{}\infty}\LL(w_t)=0$,
    \item $\lim_{t \xrightarrow{}\infty}\norm{w_t}=\infty$,
    \item $\forall i: \ \ \lim_{t \xrightarrow{}\infty} y_iw_t^Tx_i=\infty$,
\end{enumerate}
\end{lemma}

\begin{proof} As mentioned we use the exponential loss function: $\ell:u \mapsto e^{-u}$, which.\\
Since $\D$ is linearly separable, $\exists w_*$ such that $ w_*^T x_i > 0, \forall i$. Then for $w \in \R^d$:
\[
w_*^T\nabla \LL(w) = \sum_{i=1}^{n}  \underbrace{-exp(-y_i w^T x_i)}_{<0}  \underbrace{y_i w_*^T x_i}_{>0} < 0.
\]
Therefore there is no finite critical points $w$, for which $\nabla \LL(w)=0$. But gradient descent on a smooth loss with an appropriate learning rate is always guaranteed to converge to a critical point : in other words $\nabla \LL(w_t)\xrightarrow{}0$. This necessarily implies that $\norm{w_t}\xrightarrow{}\infty$, which is (2). It also implies  that $\exists t_0$ s.t, $\forall t>t_0, \forall i: y_i w_t^T x_i>0$ in order to make the exponential term converge to zero, this is (3). But in that case, we also have $\LL(w_t)\xrightarrow{}0$, which is (1).
\end{proof}

The norm of the previous solution diverges, but we can normalize it to have norm 1.

\vspace{.4cm}
\begin{theorem}
Let $\D = \{(x_i, y_i)\}_{i=1}^{n}$ be a linearly separable dataset where $\forall i \in [\![ 1, n]\!]$, $(x_i, y_i) \in \R^d\times\{-1,1\}$ and $\ell:\R \mapsto \R^*_+$ a loss function with under assumptions 1, 2 and 3.
Let $w_t$ be the iterates of gradient descent using a learning rate $\eta$ such that $0<\eta<\frac{2}{\beta\sigma^2_{max}(X)}$ and any starting point $w_0$. Then we have:
\[
\lim_{t \xrightarrow{}\infty}\frac{w_t}{\norm{w_t}}=\frac{w_{svm}}{\norm{w_{svm}}}
\]
where $w_{svm}$ is the solution to the hard margin SVM:
\[
w_{svm} = \argmin_{w\in\R^d}\norm{w}^2\ \  s.t.\ \  y_i w^T x_i\geq 1, \forall i.
\]
\end{theorem}

\begin{proof_sketch}
We will just give the main ideas behind the proof of this theorem using the exponential loss function. We will furthermore assume that $\nicefrac{w_t}{\norm{w_t}}$ converges to some limit $w_{\infty}$. For a detailed proof and in the more general case of the loss function having properties 1 to 3 please refer to \cite{soudry_implicit_2018}.

By Lemma \ref{lemma:exploding_norm} we have $\forall i:\ \lim_{t \xrightarrow{}\infty} y_iw_t^Tx_i=\infty$. As $\frac{w_t}{\norm{w_t}}$ converges to $w_{\infty}$ we can write $w_t = g(t)w_{\infty}+\rho(t)$ such that $g(t) \xrightarrow{}\infty$, $\forall i:\ y_iw^T_{\infty}x_i >0$ and $\ \lim_{t \xrightarrow{}\infty} \frac{\rho(t)}{g(t)}=0$. The gradient can then be written as:
\begin{equation}
\label{eq:neg_grad}
- \nabla \LL(w_t) = \sum_{i=1}^{n} e^{-y_iw_t^Tx_i}x_i
                = \sum_{i=1}^{n} e^{-g(t)y_iw_{\infty}^Tx_i}\ e^{-y_i\rho(t)^Tx_i}x_i
\end{equation}
We can see that as $g(t) \xrightarrow{}\infty$ only the samples with largest exponents in the sum of the right hand side of (\ref{eq:neg_grad}) will contribute to the gradient. But the exponents are maximized for  $i \in \mathcal S = argmin_i\ y_iw_{\infty}^Tx_i$ which correspond to the samples minimizing the margin: i.e. the support vectors $X_S = \{x_i, i \in \mathcal S\}$.
The negative gradient $- \nabla \LL(w_t)$ would then asymptotically become a non-negative linear combination of support vectors and because $\norm{w_t}\xrightarrow{}\infty$ (by Lemma \ref{lemma:exploding_norm}) the first gradient steps will be negligible and the limit $w_{\infty}$ will get closer and closer to a non-negative linear combination of support vectors and so will its scaled version $\hat w = w_{\infty}/\min_i y_iw_{\infty}^Tx_i$ (the scaling is done to make the margin of the support vectors equal to 1).
We can therefore write:
\begin{equation}
\hat w = \sum_{i=1}^n \alpha_ix_i\quad with\ 
\left\{
    \begin{array}{ll}
        \alpha_ix_i\geq 0 \ and\ y_i\hat w^T x_i=1\ if\ i\in \mathcal S\\
       \alpha_ix_i= 0 \ and\ y_i\hat w^T x_i>1\ if\ i\notin \mathcal S
    \end{array}
\right.
\end{equation}
We can recognize the KKT conditions for the hard margin SVM problem (see \cite{bishop_pattern_2006} Chapter 7, Section 7.1) and conclude that $\hat w = w_{svm}$. Then $\displaystyle \frac{w_{\infty}}{\norm{w_{\infty}}}=\frac{w_{svm}}{\norm{w_{svm}}}$.
\end{proof_sketch}

\begin{remark}
In the proof of Lemma \ref{lemma:exploding_norm} we have seen that $\LL(w_t)\xrightarrow{}0$. That means that gradient descent converges to a global minimum.
\end{remark}

\begin{remark}
Gradient descent has been suspected to induce a bias towards simple solutions, not only in the previous linear settings, but in deep learning as well, greatly improving generalization performance. It would explain the double descent behavior of deep learning architectures, and recent works such as \cite{Gissin2019} have been studying the learning dynamics in more complex settings.
\end{remark}
\newpage
\section{The reasons behind double descent}
\label{section:3}
In this section, we consider two settings where double descent can be empirically observed and mathematically justified, in order to give the reader some intuition on the role of inductive biases. We conclude with some references to recent related works studying optimization in the over-parameterized regime, or linking the double descent to a physical phenomenon named \emph{jamming}.

Fully understanding the mechanisms behind this phenomenon in deep learning remains an open question but, as introduced in section \ref{section:inductive biases}, inductive biases seem to play a key role.

\begin{figure}[ht]
    \begin{subfigure}{0.24 \textwidth}
        \centering
        \includegraphics[width=\linewidth]{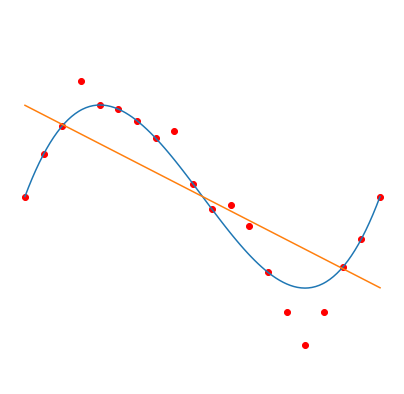}
        \caption{$d=1$}
    \end{subfigure}
    \begin{subfigure}{0.24 \textwidth}
        \centering
        \includegraphics[width=\linewidth]{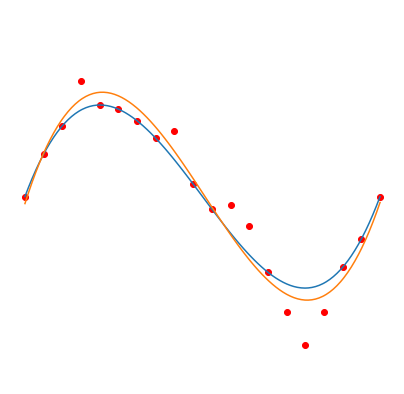}
        \caption{$d=3$}
    \end{subfigure}
    \begin{subfigure}{0.24 \textwidth}
        \centering
        \includegraphics[width=\linewidth]{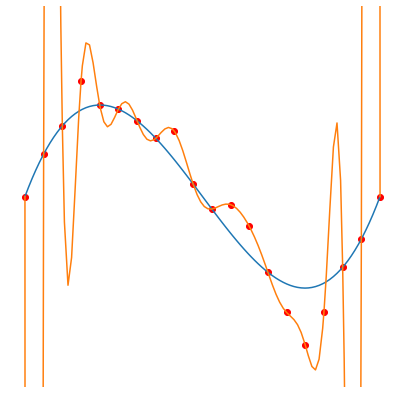}
        \caption{$d=20$}
        \label{fig:overfitting_d20}
    \end{subfigure}
    \begin{subfigure}{0.24 \textwidth}
        \centering
        \includegraphics[width=\linewidth]{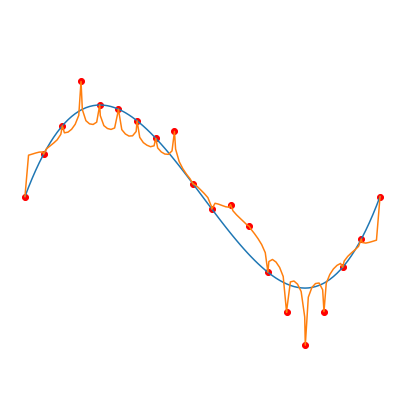}
        \caption{$d=1000$}
        \label{fig:overfitting_d1000}
    \end{subfigure}
    \caption{Fitting degree $d$ Legendre polynomials (orange curve) to $n=20$ noisy samples (red dots), from a polynomial of degree 3 (blue curve). Gradient descent is used to minimize the squared error, which leads to the smallest norm solution (considering the norm of the vector of coefficients). Taken from \cite{blog_double_descent}.}
\end{figure}

In the over-parameterized regime, empirical risk minimizers are able to interpolate the data. Intuitively :
\begin{itemize}
    \item Near the interpolation point, there are very few solutions that fit the training data perfectly. Hence, any noise in the data or model mis-specification will destroy the global structure of the model, leading to an irregular solution that generalizes badly (figure \ref{fig:overfitting_d20}). 
    \item As effective model capacity grows, many more interpolating solutions exist, including some that generalize better and can be selected thanks to the right inductive bias, e.g. smaller norm (figure \ref{fig:overfitting_d1000}), or ensemble methods.
\end{itemize}

\vspace{.4cm}

\subsection{Linear Regression with Gaussian Noise}
In this section we consider the family class $(\mathcal{H}_p)_{p\in\llbracket1,d\rrbracket}$ of linear functions $h:\R^d\mapsto \R$ where exactly $p$ components are non-zero ($1\leq p\leq d$). We will study the generalization error obtain using ERM when increasing p (which is regarded as the class complexity).
The class of predictors $\mathcal{H}_p$ is defined as follow:
\begin{definition}
For $p \in \llbracket1,d\rrbracket$, $\mathcal{H}_p$ is the set of functions $h:\R^d\mapsto \R$ of the form:
$$
h(u)=u^Tw,\quad \text{for }u \in \R^d
$$
With $w \in \R^d$ having exactly $p$ non-zero elements.
\end{definition}

Let $(X, \mathbf{\epsilon})\in \R^d\times\R$ be independent random variables with $X \sim \N(0,I)$  and $\mathbf{\epsilon} \sim \N(0,\sigma^2)$. Let $h^* \in \mathcal{H}_d$ and define the random variable $Y=h^*(X)+\sigma \mathbf{\epsilon}=X^Tw+\sigma \mathbf{\epsilon}$, with $\sigma>0$ with $w \in \R^d$ defined by $h^*$. We consider $(X_i, Y_i)_{i=1}^n$ $n$ iid copies of $(X,Y)$.
We are interested in the following problem:
\begin{equation}
\label{eq:linear_gaussian}
   \min_{h\in \mathcal{H}_d}\E[(h(X) - Y)^2]
\end{equation}

Let $\mathbf{X}\in \R^{n\times d}$ the random matrix which rows are the $X_i^T$ and  $\mathbf{Y} =(Y_1,.., Y_n)^T \in \R^n$. In the following we will assume that $\bm{X}$ is full row rank and that $n \ll d$. Applying empirical risk minimization we can write:
\begin{equation}
\label{eq:linear_gaussian_erm}
\min_{w\in \R^d} \frac{1}{2}\norm{\bm{X} w - \bm{Y}}^2
\end{equation}

\vspace{.4cm}
\footnotetext[1]{
The notation used for the random p-submatrix and random p-subvector is not common and is introduced for clarity.}
\begin{definition}[Random p-submatrix/p-subvector\footnotemark]
For any $(p,q) \in \llbracket 1, d\rrbracket^2$ such that $p+q=d$ and matrix $\bm{A} \in \R^{n\times d}$ and column vector $v\in \R^d$, we will denote by $\bm{\Ap}$ (resp. $\vp$) the sub-matrix (resp. sub-vector) obtained by randomly selecting a subset of p columns (resp. elements), and by $\bm{\Aq} \in \R^{n\times q}$ and $\vq\in \R^{q}$ their discarded counterpart.
\end{definition}

In order to solve (\ref{eq:linear_gaussian_erm}) we will search for a solution in $\mathcal{H}_p \subset \mathcal{H}_d$ and increase $p$ progressively which is a form of structural empirical risk minimization as $\mathcal{H}_p \subset \mathcal{H}_{p+1}$ for any $p<d$.

Let $p \in \llbracket 1, d\rrbracket$, we are then interested in the following sub-problem:
$$
\min_{w\in \R^p} \frac{1}{2}\norm{\bm{\Xp} w - y}^2
$$
We have seen in proposition \ref{prop:smalestnorm} of section \ref{section:gd_least_square} that the least norm solution is $\p{\hat w}=\bm{\Xp}^+y$. If we define $\q{\hat w} := 0$ then we will consider as a solution of the global problem (\ref{eq:linear_gaussian}) ${\hat w:=\phi_p(\p{\hat w},\q{\hat w})}$ where $\phi_p: \R^p\times\R^{q}\mapsto \R^d$ is a map rearranging the terms of $\p{\hat w}$ and $\q{\hat w}$ to match the initial indices of $w$.

\vspace{.4cm}
\begin{theorem}
\label{thm:double_descent_lr}
Let $(x, \epsilon)\in \R^d\times\R$ independent random variables with $x \sim \N(0,I)$  and $\epsilon \sim \N(0,\sigma^2)$, and $w \in \R^d$. we assume that the response variable $y$ is defined as $y=x^Tw +\sigma \epsilon$. Let $(p,q) \in \llbracket 1, d\rrbracket^2$ such that $p+q=d$, $\bm{\Xp}$ the randomly selected $p$ columns sub-matrix of X. Defining $\hat w:=\phi_p(\p{\hat w},\q{\hat w})$ with $\p{\hat w}=\bm{\Xp}^+y$ and $\q{\hat w} = 0$.\\
The risk of the predictor associated to $\hat w$ is:
$$
\E[(y-x^T\hat w)^2] = 
\begin{cases}
(\norm{\wq}^2+\sigma^2)(1+\frac{p}{n-p-1}) &\quad\text{if } p\leq  n-2\\
+\infty &\quad\text{if }n-1 \leq p\leq  n+1\\
\norm{\wp}^2(1-\frac{n}{p}) +  (\norm{\wq}^2+\sigma^2)(1+\frac{n}{p-n-1}) &\quad\text{if }p\geq n+2\end{cases}
$$
\end{theorem}

\begin{proof}
Because $x$ is zero mean and identity covariance matrix, and because $x$ and $\epsilon$ are independent:

\begin{align*}
\E[(y-x^T\hat w)^2] = \E[(x^T(w-\hat w) + \sigma \epsilon)^2] &= \sigma^2 + \E[\norm{w-\hat w}^2]\\
&= \sigma^2 + \E[\norm{\wp-\p{\hat w}}^2]+\E[\norm{\wq-\q{\hat w}}^2]
\end{align*}

and because $\q{\hat w}=0$, we have: 
$
\E[(y-x^T\hat w)^2] =  \sigma^2 + \E[\norm{\wp-\p{\hat w}}^2]+\norm{\wq}^2
$

The classical regime ($p\leq n$) as been treated in \cite{breiman_how_1983}. We will then consider the interpolating regime ($p \geq n$). Recall that X is assumed to be of rank $n$. Let $\eta = y - \bm{\Xp} \wp$. 
We can write :
\begin{equation*}
    \begin{aligned}
    \wp-\p{\hat w} = \wp - \bm{\Xp}^+y =
    \wp - \bm{\Xp}^+(\eta + \bm{\Xp} \wp)
        = (\bm{I}- \bm{\Xp}^+\bm{\Xp})\wp - \bm{\Xp}^+ \eta
    \end{aligned}
\end{equation*}

It is easy to show (left as an exercise) that $(\bm{I}- \bm{\Xp}^+\bm{\Xp})$ is the matrix of the orthogonal projection on $\text{Ker}(\bm{\Xp})$. Furthermore, $-\bm{\Xp}^+ \eta \in \text{Im}(\bm{\Xp}^+)=\text{Im}(\bm{\Xp}^T)$.
Then ${(\bm{I}- \bm{\Xp}^+\bm{\Xp})\wp}$ and $-\bm{\Xp}^+ \eta$ are orthogonal and the Pythagorean theorem gives:
\[
\norm{\wp-\p{\hat w}}^2 = \norm{(\bm{I}- \bm{\Xp}^+\bm{\Xp})\wp}^2 + \norm{\bm{\Xp}^+ \eta}^2
\]

We will treat each term of the right hand side of this equality separately.

\begin{itemize}
    \item $ \norm{(\bm{I}- \bm{\Xp}^+\bm{\Xp})\wp}^2$: $\bm{\Xp}^+\bm{\Xp}$ is the matrix of the orthogonal projection on $\text{Im}(\bm{\Xp}^T)=\text{Im}(\bm{\Xp}^+)$, then using again the Pythagorean theorem gives:
\[
\norm{(\bm{I}- \bm{\Xp}^+\bm{\Xp})\wp}^2 = \norm{\wp}^2 - \norm{\bm{\Xp}^+\bm{\Xp}\wp}^2
\]

Because $\bm{\Xp}^+\bm{\Xp}$ is the matrix of the orthogonal projection on $\text{Im}(\bm{\Xp}^T)$ we can write $\bm{\Xp}^+\bm{\Xp}\wp$ as a linear combination of rows of $X_p$, then using the fact that the $x_i$ are i.i.d and of standard normal distribution we have:
\[
\E[\norm{\bm{\Xp}^+\bm{\Xp}\wp}^2]=\norm{\wp}^2\frac{n}{p}\quad \text{ then }\quad \E[\norm{(\bm{I}- \bm{\Xp}^+\bm{\Xp})\wp}^2]=\norm{\wp}^2(1-\frac{n}{p})
\]
    \item $ \norm{\bm{\Xp}^+ \eta}^2$: The calculation of this term used the "trace trick" and the notion of distribution of inverse-Wishart for pseudo-inverse matrices and is beyond the scope of this tutorial. It can be shown that:
    $$\E[\norm{\bm{\Xp}^+ \eta}^2]= \begin{cases}
(\norm{\wq}^2+\sigma^2)(\frac{n}{p-n-1}) &\quad\text{if } p\geq  n+2\\
+\infty &\quad\text{if }p \in \{n,n+1\}
\end{cases}
$$
 
\end{itemize}

\end{proof}

\begin{corollary}
Let $T$ be a uniformly random subset of $\llbracket 1, d\rrbracket$ of cardinality p.
Under the setting of Theorem \ref{thm:double_descent_lr} and taking the expectation with respect to $T$. The risk of the predictor associated to $\hat w$ is:
$$
\E[(Y-X^T\hat w)^2] = 
\begin{cases}
\left((1-\frac{p}{d})\norm{w}^2+\sigma^2\right)(1+\frac{p}{n-p-1}) &\quad\text{if } p\leq  n-2\\
\norm{w}^2\left(1-\frac{n}{d}(2- \frac{d-n-1}{p-n-1})\right) +\sigma^2(1+\frac{n}{p-n-1}) &\quad\text{if }p\geq n+2\end{cases}
$$
\end{corollary}

\begin{proof}
Since T is a uniformly random subset of  $\llbracket 1, d\rrbracket$ of cardinality p:
$$
\E[\norm{\wp}^2] = \E[\sum_{i\in T}w_i^2]= \E[\sum_{i=1}^{d}w_i^2 \1_{T}(i) ]=\sum_{i=1}^{d}w_i^2 \E[\1_{T}(i) ]=\sum_{i=1}^{d}w_i^2 \mathbb{P}[i \in T]=\norm{w}^2 \frac{p}{d}
$$
and, similarly:
$$
\E[\norm{\wq}^2] =\norm{w}^2 (1-\frac{p}{d})
$$
Plugging into Theorem \ref{thm:double_descent_lr} ends the proof.
\end{proof}

\subsection{Random Fourier Features}
\label{section:RFF}
In this section we consider the RFF model family \cite{Rahimi2009} as our class of predictors $\mathcal{H}_N$.

\begin{definition}
We call \emph{Random Fourier Features (RFF)} model any function $h: \mathbb{R}^d \rightarrow \mathbb{R}$ of the form :
$$
h(x) = \beta^T z(x)
$$
With
$z(x) = \sqrt{\frac{2}{N}} \begin{bmatrix}\cos(\omega_1^T x + b_1) \\ \vdots \\ \cos(\omega_N^T x + b_N)\end{bmatrix}$,
$\beta \in \mathbb{R}^N$ the parameters of the model
and $\forall i \in \llbracket 1,N \rrbracket \begin{cases}\omega_i \sim \mathcal{N}(0, \sigma^2 I_d) \\ b_i \sim \mathcal{U}([0, 2\pi])\end{cases}$. The vectors $\omega_i$ and the scalars $b_i$ are sampled before fitting the model, and $z$ is called a \emph{randomized map}.
\end{definition}

The RFF family is a popular class of models that are linear w.r.t. the parameters $\beta$ but non-linear w.r.t. the input $x$, and can be seen as two-layer neural networks with fixed weights in the first layer. In a classification setting, using these models with the hinge loss amounts to fitting a linear SVM to $n$ feature vectors (of dimension $N$). RFF models are typically used to approximate the Gaussian kernel and reduce the computational cost when $N \ll n$ (e.g. kernel ridge regression when using the squared loss and a $l_2$ regularization term). In our case however, we will go beyond $N=n$ to observe the double descent phenomenon.

\vspace{.4cm}
\begin{remark}
Clearly, we have $\mathcal{H}_N \subset \mathcal{H}_{N+1}$ for any $N \geq 0$.
\end{remark}

\vspace{.4cm}
\begin{proposition}[Approximation of the Gaussian Kernel, informal]
Let $k:(x,y) \rightarrow e^{-\frac{1}{2\sigma^2}||x-y||^2}$ be the Gaussian kernel on $\mathbb{R}^d$, and let $\mathcal{H}_{\infty}$ be a class of predictors where empirical risk minimizers on $\mathcal{D}_n = \{(x_1, y_1), \dots, (x_n, y_n)\}$ can be expressed as $h: x \rightarrow \sum_{k=1}^n \alpha_k k(x_k, x)$. Then, as $N \rightarrow \infty$, $\mathcal{H}_N$ becomes a closer and closer approximation of $\mathcal{H}_{\infty}$.
\end{proposition}

\begin{proof_sketch}
For any $x, y \in \mathbb{R}^d$, with the vectors $\omega_k \in \mathbb{R}^d$ sampled from $\mathcal{N}(0, \sigma^2 I_d)$ :
\begin{equation*}
\begin{aligned}
k(x,y)  &= e^{-\frac{1}{2\sigma^2}(x-y)^T(x-y)} 
        \overset{(1)}{=} \mathbb{E}_{\omega \sim \mathcal{N}(0, \sigma^2 I_d)}[e^{i \omega^T(x-y)}]
        \overset{(2)}{=} \mathbb{E}_{\omega \sim \mathcal{N}(0, \sigma^2 I_d)}[\cos(\omega^T(x-y))]\\
        &\approx \frac{1}{N} \sum_{k=1}^N \cos(\omega_k^T(x-y)) 
        = \frac{1}{N} \sum_{k=1}^N 2 \cos(\omega_k^T x + b_k) \cos(\omega_k^T y + b_k)
        \overset{(3)}{=} z(x)^T z(y)
\end{aligned}
\end{equation*}
Where $(1)$ and $(3)$ are left as an exercise, with indications in \cite{blog_rff} if needed, and $(2)$ is because $k(x,y) \in \mathbb{R}.$

Hence, for $h \in \mathcal{H}_{\infty}$ :
$h(x) 
= \sum_{k=1}^n \alpha_k k(x_n, x)
\approx \underbrace{\left(\sum_{k=1}^N \alpha_k z(x_k) \right)^T}_{\beta} z(x)
$
\end{proof_sketch}

A complete definition is outside of the scope of this tutorial, but $\mathcal{H}_{\infty}$ is actually the \emph{Reproducing Kernel Hilbert Space (RKHS)} corresponding to the Gaussian kernel, for which RFF models are a good approximation when sampling the random vectors $\omega_i$ from a normal distribution.

\begin{figure}[ht]
    \begin{subfigure}{0.45 \textwidth}
        \centering
        \includegraphics[width=0.9\textwidth, height=4cm]{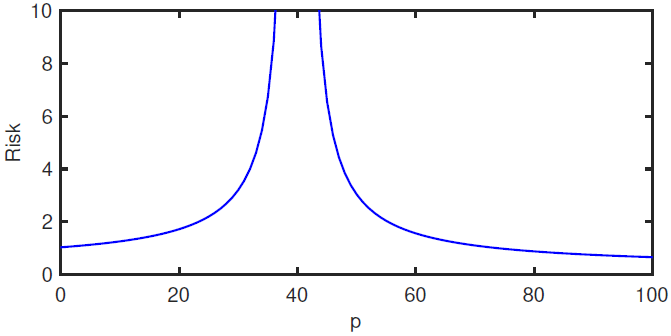}
        \vspace{2mm}
        \caption{Plot of risk $\E[(y-x^T\hat{w})^2]$ as a function of $p$, under the random selection model of the subset of $p$ features. Here $\norm{w}^2=1$,  $\sigma^2=1/25$, $d=100$ and $n=40$. Taken from \cite{belkin_two_2020}}
        \label{fig:doubledescent_gaussian}
    \end{subfigure}
    \begin{subfigure}{0.45 \textwidth}
        \centering
    \vspace{9mm}
        \includegraphics[width=\textwidth,height=4.5cm]{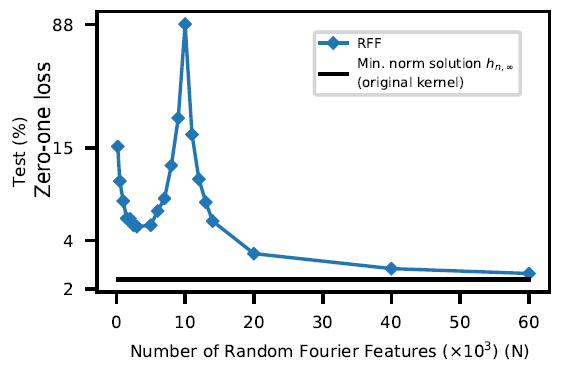}
        \caption{Model-wise double descent risk curve for RFF model on a subset of MNIST ($n=10^4$, 10 classes), \emph{choosing the smallest norm predictor $h_{n,N}$} when $N > n$. The interpolation threshold is achieved at $N=10^4$. Taken from \cite{belkin_reconciling_2019}, which uses an equivalent but slightly different definition of RFF models.}
        \label{fig:rff_mnist}
    \end{subfigure}
    \caption{Risk curves as a function of model capacity.}
\end{figure}

We use ERM to find the predictor $h_{n,_N} \in \mathcal{H}_N$ and, in the interpolation regime where multiple minimizers exist, we choose the one whose parameters $\beta \in \mathbb{R}^N$ have the smallest $l_2$ norm. This training procedure allows us to observe a model-wise double descent (figure \ref{fig:rff_mnist}). Indeed, in the under-parameterized regime, statistical analyses suggest choosing $N \propto \sqrt{n} \log(n)$ for good test risk guarantees \cite{Rahimi2009}. And as we approach the interpolation point (around $N = n$), we observe that the test risk increases then decreases again.

In the over-parameterized regime ($N \geq n$), multiple predictors are able to fit perfectly the training data. As $\mathcal{H}_N \in \mathcal{H}_{N+1}$, increasing $N$ leads to richer model classes and allows constructing interpolating predictors that are more regular, with smaller norm (eventually converging to $h_{n,\infty}$ obtained from $\mathcal{H}_{\infty}$). As detailed in theorem \ref{thm:rff_bound} (in a noiseless setting), the small norm inductive bias is indeed powerful to ensure small generalization error. 

\vspace{.4cm}
\begin{theorem}[Belkin et al. \cite{belkin_reconciling_2019}]
\label{thm:rff_bound}
Fix any $h^* \in \mathcal{H}_{\infty}$. Let $(X_1, Y_1), \dots ,(X_n, Y_n)$ be i.i.d. random variables, where $X_i$ is drawn uniformly at random from a compact cube $\Omega \in \mathbb{R}^d$, and $Y_i = h^*(X_i)$ for all $i$. There exists constants $A,B > 0$ such that, for any interpolating $h \in \mathcal{H}_{\infty}$ (i.e., $h(X_i) = Y_i$ for all $i$), so that with high probability :
$$
\sup_{x \in \Omega}|h(x) - h^*(x)| < A e^{-B(n/log n)^{1/d}} 
(||h^*||_{\mathcal{H}_{\infty}} + ||h||_{\mathcal{H}_{\infty}})
$$
\end{theorem}

\begin{proof}
We refer the reader directly to \cite{belkin_reconciling_2019} for the proof.
\end{proof}

\newpage
\subsection{Related works}
\label{section:related-works}
\subsubsection{Optimization in the over-parameterized regime}
For reasons that are still under investigation, overparameterization seems beneficial not only in the statistical learning framework, but from an optimization standpoint as well as it facilitates convergence to global minima, in particular with the gradient descent procedures.

The optimization problem can be framed as minimizing a certain loss function $\LL(w)$ with respect to its parameters $w \in \mathbb{R}^N$, such as the square loss $\LL(w) = \frac{1}{2} \sum_{i=1}^n (f(x_i, w) - y_i)^2$ where $\{(x_i, y_i)\}_{i=1}^{n}$ is our given training dataset and $f : (\mathbb{R}^d \times \mathbb{R}^N) \rightarrow \mathbb{R}$ is our model.

\begin{exercise}
Assume that $\ell: \mathcal{Y} \rightarrow \mathbb{R}$ is convex and $f : \mathcal{X} \rightarrow \mathcal{Y}$ is linear. Show that $\ell \circ f$ is convex.
\end{exercise}

When $f$ is non-linear however (which is habitually the case in deep learning) the landscape of the loss function is generally non-convex. Therefore, first order methods such as GD or SGD are likely to converge and get trapped in spurious local minima, depending on the initialization. Yet, in the over-parameterized regime where there are multiple global minima interpolating almost perfectly the data, it seems that SGD has no problem converging to these solutions, despite the highly non-convex setting. Recent works are trying to explain this phenomenon. 

For instance, \cite{Oymak2019} shows that, for one-hidden layer neural networks that (1) have smooth activation functions, (2) are over-parameterized, i.e. $N \geq C n^2$ where C depends on the distribution of the data and (3) are initialized with i.i.d. $\mathcal{N}(0,1)$ entries, then with high probability GD converges quickly to a global optimum. Similar results also hold for ReLU activation functions and for SGD. 

\begin{figure}[ht]
\centering
    \centering
    \includegraphics[scale=0.3]{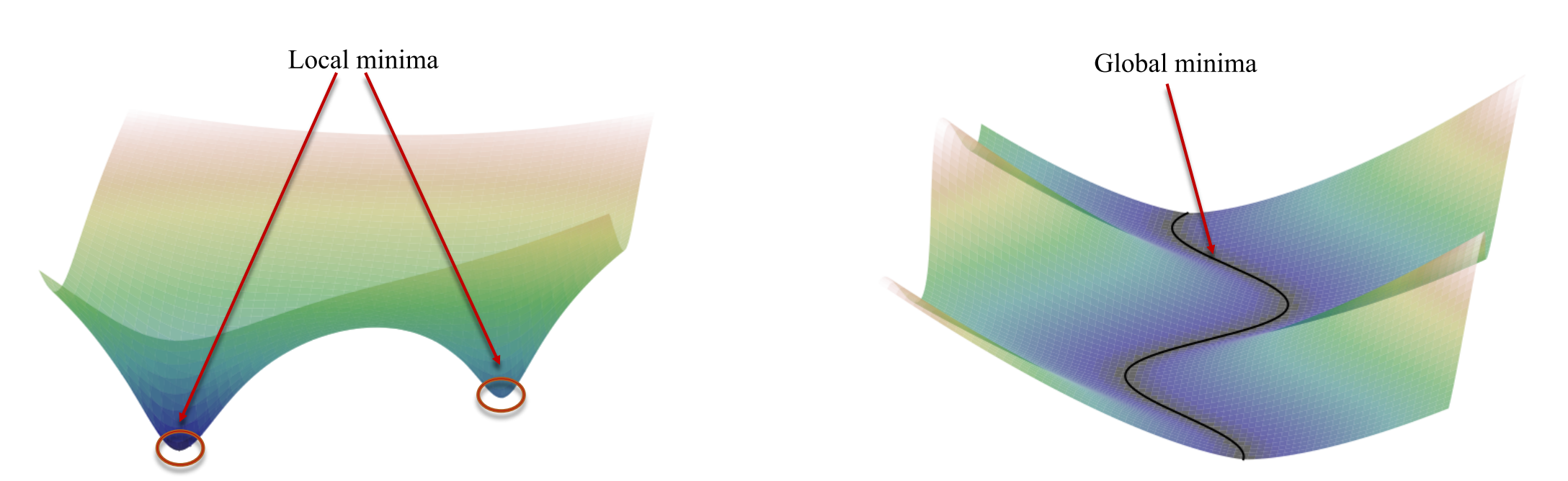}
    \caption{\emph{Left :} Loss landscape of under-parameterized models, locally convex at local minima. \emph{Right :} Loss landscape of over-parameterized models, incompatible with local convexity. Taken from \cite{Liu2020}.}
    \label{fig:loss-landscape}
\end{figure}

In \cite{Liu2020}, the authors show that sufficiently over-parameterized systems, including wide neural networks, generally satisfy a condition that allows gradient descent to converge efficiently, for a broad class of problems. They use the PL-condition (from Polyak and Lojasiewicz \cite{polyak1963gradient}) which does not require convexity but is sufficient for efficient minimization by GD. One key point is that the loss function $\LL(w)$ is generally non-convex in the neighborhood of minimizers. Due to the over-parameterization, the Hessian matrices $\nabla^2 \LL(w)$ are positive semi-definite but not positive definite in these neighborhoods, which is incompatible with convexity for non-linear sets of solutions. This is in contrast to the under-parameterized landscape which generally has multiple isolated local minima with positive definite Hessian matrices. Figure \ref{fig:loss-landscape} illustrates this.

In addition to better convergence guarantees, over-parameterization can even accelerate optimization. By working with \emph{linear} neural networks (hence fixed expressiveness), \cite{Arora2018} finds that increasing depth has an implicit effect on gradient descent, combining certain forms of \emph{momentum} and \emph{adaptive learning rates} (two well-known tools in the field of optimization). They observe the acceleration for non-linear networks as well (replacing weight matrices by a product of matrices, for fixed expressiveness), and even when using explicit acceleration methods such as Adam.

\subsubsection{Neural networks as a physical system : the jamming transition}
In order to study the loss landscape, \cite{spigler_jamming_2019} make an analogy between neural networks and complex physical systems with non-convex energy landscape, called glassy systems. Indeed, the loss function can be interpreted as the potential energy of the system $f$, with a large number of parameters $N$ (degrees of freedom). By considering the hinge loss, the minimization of $\LL(w;\Dn)$ actually amounts to a constraint-classification problem (with $n$ constraints, $N$ continuous degrees of freedom), already studied in physics.

Using this analogy, they show that the behavior of deep networks near the interpolation point is similar to the behavior of some granular systems, that undergo a critical \emph{jamming transition} when their density increases such that they are forced to be in contact one another. In the under-parameterized regime, not all the training examples can be classified correctly, which leads to unsatisfied constraints. But in the over-parameterized regime, there is no stable local minima : the network reaches a global minima zero training loss.

\begin{figure}[H]
\centering
    \centering
    \includegraphics[scale=0.2]{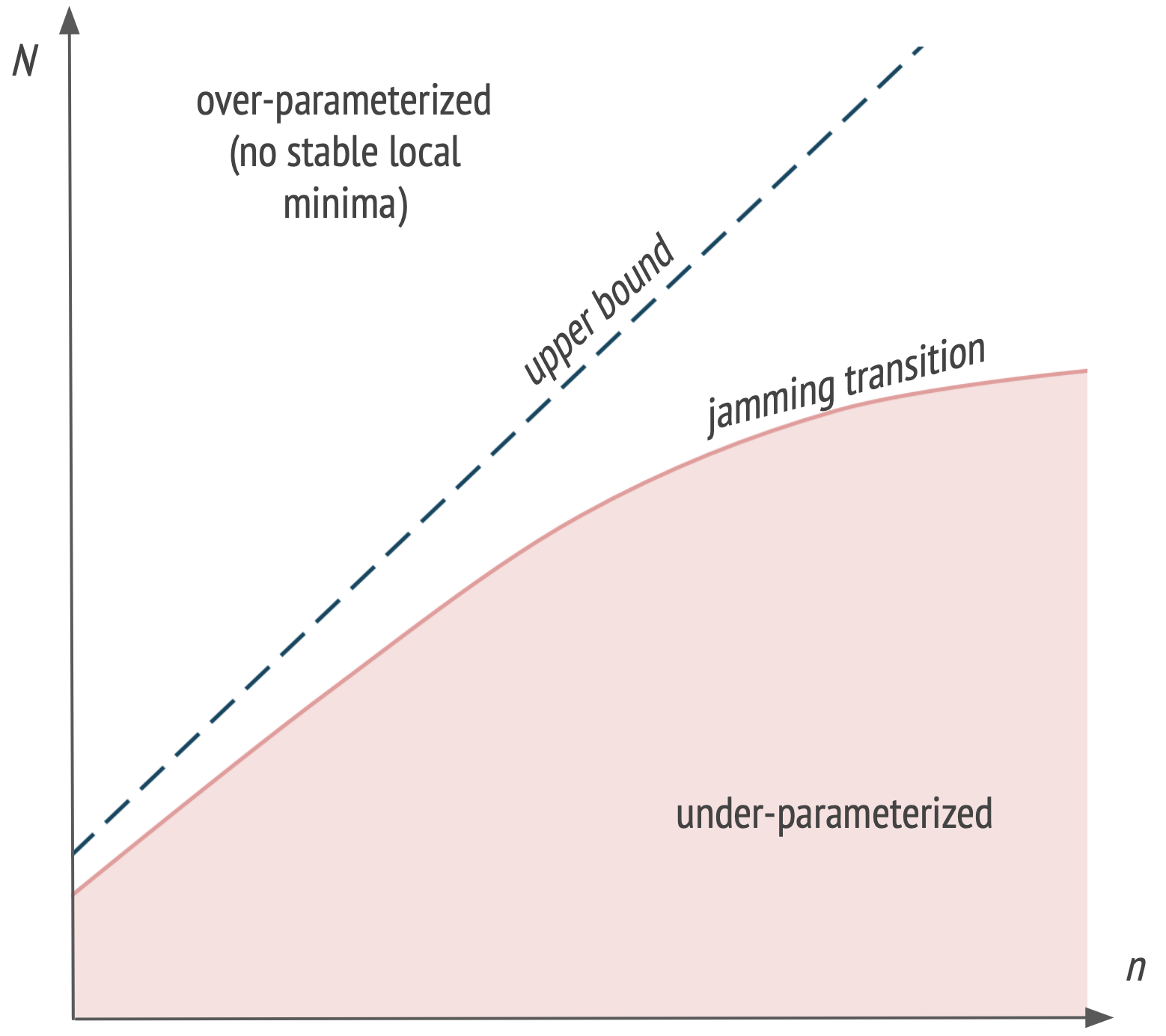}
    \caption{$N$ : degrees of freedom, $n$ : training examples. Inspired from \cite{spigler_jamming_2019}}
    \label{fig:jamming-transition}
\end{figure}

As illustrated in figure \ref{fig:jamming-transition}, the authors are able to quantify the location of the jamming transition in the $(n, N)$ plane (considering $N$ as the \emph{effective} number of parameters of the network). Considering a fully-connected network with arbitrary depth, ReLU activation functions and a dataset of size $n$, they give a linear upper bound on the critical number of parameters $N^*$ characterizing the jamming transition : $N^* \leq \frac{1}{C_0} n$ where $C_0$ is a constant. In their experiments, it seems that the bound is tight for random data but that $N^*$ increases sub-linearly with $n$ for structured data (e.g. MNIST), as illustrated on figure \ref{fig:jamming-transition}.

Similarly to other works, they observe a peak in test error at the jamming transition. In \cite{Geiger2020}, using the same setting of fixed-depth fully-connected networks, they argue that this may be due to $||f||$ diverging near the interpolation point $N=N^*$. Interestingly, they also observe that near-optimal generalization can be obtained using an ensemble average of networks with $N$ slightly beyond $N^*$.

\section{Conclusion}
From a statistical learning point of view, deep learning is a challenging setting to study and some recent empirical successes are not yet well understood. The double descent phenomenon, arising from well-chosen inductive biases in the over-parameterized regime, has been studied in linear settings and observed with deep networks \cite{nakkiran_deep_2019}. 

In addition to the references presented in section \ref{section:related-works}, other lines of work seem promising. Notably, \cite{Gissin2019, Neyshabur2015, soudry_implicit_2018, gunasekar_implicit_2017} are working towards a better understanding of the implicit bias induced by optimization algorithms. Finally, we refer the reader to subsequent works of Belkin \textit{et al.} such as \cite{Chen2020}, that finds \emph{multiple descent} curves with an arbitrary number of peaks, due to the interaction between the properties of the data and the inductive biases of learning algorithms.

\bibliographystyle{abbrvnat}
\footnotesize
\bibliography{biblio}

\end{document}